\documentclass{article}

\usepackage{PRIMEarxiv}

\usepackage[utf8]{inputenc} % allow utf-8 input
\usepackage[T1]{fontenc}    % use 8-bit T1 fonts
\usepackage{hyperref}       % hyperlinks
\usepackage{url}            % simple URL typesetting
\usepackage{booktabs}       % professional-quality tables
\usepackage{amsfonts}       % blackboard math symbols
\usepackage{nicefrac}       % compact symbols for 1/2, etc.
\usepackage{microtype}      % microtypography
\usepackage{lipsum}
\usepackage{fancyhdr}       % header
\usepackage{graphicx}
\usepackage[numbers]{natbib}
\usepackage{arabtex}
\usepackage{utf8}
\setcode{utf8}
\usepackage{multirow}
\usepackage{multicol}
\graphicspath{{media/}}     % organize your images and other figures under media/ folder

\title{STF: SENTENCE TRANSFORMER FINE-TUNING FOR TOPIC
CATEGORIZATION WITH LIMITED DATA
\thanks{\textit{\underline{Citation}}: 
\textbf{Authors. Title. Pages.... DOI:000000/11111.}} 
}

\author{
  Kheir Eddine Daouadi, Yaakoub Boualleg, Oussama Guehairia \\
  Laboratory of Vision and Artificial Intelligence (LAVIA) \\
  Echahid Cheikh Larbi Tebessi University \\
  Tebessa, Algeria\\
  \texttt{\{kheireddine.daouadi, yaakoub.boualleg, oussama.guehairia\}@univ-tebbesa.dz} \\
}

\begin{document}
\maketitle
Nowadays, topic classification from tweets attracts several researchers’ attention. Different classification systems have been suggested thanks to these research efforts. Nevertheless, they are confronted major challenge owing to the low performance metrics because of the limited labeled data. We propose, Sentence Transformers Fine-tuning (STF), a topic detection system that leverages pre-trained Sentence Transformers models and Fine-tuning to classify topics from tweets accurately. Moreover, extensive parameter sensitivity analyses were established to fine-tune STF parameters’ for our topic classiﬁcation task to achieve the best performance results. Experiments on two benchmark datasets demonstrated that: (1) the proposed STF can be effectively used for classifying tweet topics and outperform the latest state-of-the-art approaches; (2) the proposed STF does not require a huge amount of labeled tweets to achieve good accuracy, which is the lack in the popular of the state-of-the-art approaches. Our main contribution is the achievement of promising results in tweet topic classification by applying pre-trained sentence transformers language models

% keywords can be removed
\keywords{Transfer Learning \and Sentence Transformers \and Fine-tuning \and Topic Classification \and Twitter}

\section{Introduction}

Social media have attracted the attention of politicians, researchers, companies and even governmental institutions in recent years. This gain of interest is due to the increasing applications targeted by the aforementioned actors. The emergence of social media has motivated a large number of research axes such as topic detection \cite{khatua2019tale,khairidp,oliveira2018politicians,da2,da3,yang2018using,da1}, bot detection \cite{khairibotdf,jucs2020real,dadifin} and organization detection \cite{dadifin,lpkm2018organization,daouadi2018towards}. As the growth and uses of Twitter are increasing rapidly, which make it as a source of Big Data with 500 million posts per day \footnote{\url{https://www.omnicoreagency.com/twitter-statistics/}}. Consequently, it is necessary and important to classify topics into various classes with high accuracy. The ability to categorize Tweets topics is required for developing of numerous application such as competitive intelligence detection and recommendation engines. Besides, it can help users focus on certain topic categories according to their interest.

Today, through Twitter, researchers attempt to propose approaches for topic classiﬁcation. However, they have to confronted major challenge owing to the low performance metrics because of the limited labeled data. Automatic tweet topic detection based on traditional machine learning classiﬁers like Logistic Regression (LR) and Multinomial Naïve Bayes (MNB) have shown good results. However, they are based on the handcrafted features extracted depend on some pre-deﬁned techniques like Term Frequency Inverse Document Frequency (TF-IDF) and Bag of Word (BoW) etc. Recently, Convolution neural network (CNN) and Long Short Term Memory (LSTM) have already shown good results for topic detection on Twitter. However, they are based on automatic feature extracted depend on some pre-deﬁned techniques such as the word embedding models.

Existing supervised learning classifiers show remarkable performance for tweet topic classification, but they require a huge amount of labeled training tweets. The tweets labeling process is costly and labor-intensive, while hinders the deployment of artificial intelligence systems in the industry. Transfer learning achieves substantial robustness for small data resource scenarios. Transfer learning in Natural language processing \cite{ruder2019transfer} comprises two steps: A Pre-trained Language Model trained based on unlabeled data while the second step is fine-tuning for a particular task.

In this paper, we present a comparative analysis of different machine learning techniques for classifying topic on Twitter. We evaluate the models using two datasets that contain tweets annotated for topic classification. Our experimental results show that transfer learning based models achieve the best results. The main contributions of this study can be summarized as follows:
\begin{itemize}
    \item We evaluate the three suggested deep learning architectures (LSTM, CNN and Bi-LSTM) as well as traditional machine leaning models (MNB and LR). We use them as our baselines.
    \item We evaluate and compare recent transformers and sentence transformers based pre-trained language models.
    \item We investigate the comparative of the sentence transformers models performance with our baselines systems and with the latest state-of-the-art results in tweet topic classification.
\end{itemize}
The rest of the paper is organized as follows. Section \ref{RW} presents the related works. Section \ref{DM} discusses the data and methodology. Section \ref{ER} focuses on the experiments and evaluation results. Section \ref{CC} concludes the paper with summarizing the contribution.

\section{Related Works}\label{RW}

Thanks to its importance, tweet topic classification task has drawn the attention of several researchers. In the literature, many systems and techniques have been proposed to deal with this classification problem. They follow two principal approaches: a traditional approach, and a deep learning approach.
\subsection{Traditional approaches}
In this context, researches have focused traditional learning algorithm like Multinomial Naïve Bayes (MNB) and Logistic Regression (LR) based on feature engineering. The majority of the leveraged features were lexical based features, such as Term Frequency –Inverse Document Frequency (TF-IDF) and Bag-of words BoW. Some examples of traditional approaches are briefly described in the following.

Authors in \cite{hamoud} leverage BoW to distinguish English tweets being apolitical or political. They showed that Support Vector Machine (SVM) classifier achieved Accuracy result of 88\%. 

Likewise, authors in \cite{selvaperumal} categorize English tweets based on its topics (sport, politics, news, education or entertainment). They leverage BoW, URLs in the tweet, influential users’ tweets and re-tweeted tweets. The best experimental result is achieved using SVM classifiers, yielded Accuracy result of 75\%. 
In a similar, authors in \cite{sousa} classify Portuguese tweets into apolitical or political based on BoW. They demonstrated that LR classifier outperformed deep learning ones, yielded Accuracy result of 87.5\%. 

Besides, authors in \cite{mello} leverage BoW in order to categorize Dutch tweets being apolitical and political. They demonstrated that leveraging ensemble learning is a possible solution to improve the performance metric, yielded Accuracy result of 97\%. 

Furthermore, authors in \cite{vadivukarassi} categorize English tweets based on its topics (health, sport, politics or business). They leverage TF–IDF with SVM classifier, yielded Accuracy result of 79\%. 

Likewise, authors in \cite{khatua2019tale} assume that training word2vec model based on a specific domain is better than using pre-trained ones for identifying English crisis-related tweets. They compare the performance of the Word2vec for various hyper-parameter settings and model architectures. They showed that Extra Tree classifier and Skip-gram model achieved the best experimental result, yielded Accuracy result of 91\%. 
\subsection{Deep learning approaches}
In this context, deep learning approaches based on neural networks, which can automatically learn the representation of input texts with different levels of abstraction and subsequently use the learnt knowledge to perform the classification task. The most popular deep learning architectures adopted in the field of tweet topic classification are the Long Short Term Memory (LSTM) and Convolution Neural Network (CNN). Some examples of deep leaning approaches are briefly described in the following.

In their approaches, authors in \cite{yang2018using} classify Spanish and English tweets being related to election or not. They demonstrated that Word2vec trained on general tweets is a possible solution to improve the performance metric. They explore the accuracy result of various hyper-parameter settings and model architectures. The best experimental results are achieved using CNN classifier, yielded Accuracy results of 76.5\% and 81.8\% for Spanish and English tweets, respectively. 

Besides, authors in \cite{oliveira2018politicians} classify Portuguese tweets being of apolitical and political. They leverage the Word2vec model trained on Google News with CNN classifier, yielded Accuracy result of 91\%. 

In a similar, authors in \cite{serban2019real} classify English tweets from those health related or not based on the Fasttext embedding model. The best experimental result is achieved using CNN classifier, yielded Accuracy result of 85.4\%. 

Furthermore, authors in \cite{yuan2018incorporating} assume that using weakly annotatetd tweets to pre-train tweet representations can improve the performance metric of topic classification task. The propose framework learn tweet representation based on the weakly annotated tweets and then train the classifiers based on well labeled tweets. They leveraged LSTM classifier and Word2vec trained on Google News. The proposed model, designed to categorize English tweets being non-discrimination or discrimination, yielded Accuracy results of 93.3\%. 

Likewise, authors in \cite{khairidp} incorporate contextual Word2vec, WorldNet and Deep Forest.  The contextual Word2vec model is trained based on skip-gram and well-labelled tweets, this is used in order to obtain word embedding. For word not appearing in the Word2vec model, they used Wordnet to obtain their synonyms, if one of its synonyms existed in the Word2vec model their embedding vector was used else a zero vector embedding was attributed to the word. The deep forest is used for the classification, yielded Accuracy results of 96.91\%, 98.33\% and 95.23\%.

After careful review of previous studies, we realize that a number of current pre-trained language model based on transformer have not yet been tested for tweet topic classification. Moreover, to the best of our knowledge, there are no study where transformer and sentence transformers pre-trained language models are compared to demonstrate their validity for tweet classification. In this work, we rely on transfer learning models due of its major advantages: it is able to capture long-term dependencies in language and it does not need a large dataset. In addition, we perform a comparative study between deep learning systems and traditional machine learning models considering them our baselines.
\section{Data and methodology} \label{DM}
\subsection{Dataset description}
We test our method on two benchmark English datasets composed of tweets that may discuss about different topics.

The first dataset was published by \cite{ibtihel2018semantic}, the basic characteristics of the dataset used in our experiments are shown in Table \ref{tab:tab1}. For collecting the tweets, the authors used Twitter Search API library with a list of hashtags that trigger and attract the Twitter topical content. Then, the retrieved tweets were manually annotated by two annotators to get rid of any annotator bias. The annotators were provided with a guide to follow in order to distinguish the tweets classes, yielded a dataset of 1330 labelled tweets belonging into one of the six classes namely Business, Sports, Technology, Politics, Entertainment and Education. 

The second dataset was published by \cite{magdy2015distant}, the basic characteristics of the dataset used in our experiments are shown in Table \ref{tab:tab1}. For collecting the tweets, the authors used Twitter API library to stream tweets that contain hashtag similar to class names. For instance, for the class Science\&Technology, they retrieved tweets containing \#science or \#technology hashtags. This helped authors retrieve a collection of tweets that, with high probability, had a considerable number of representatives for each class of interest. Then, for each class the authors randomly selected 200 tweets, removed the hashtag that relate them with their possible class, and submitted them for annotation. For each selected tweet, the authors asked at least three annotators whether the tweet matches the assumed class label or not. Yielded 1615 that matching the assumed class.

\begin{table*}[]
\centering
\caption{Statistics about our interested datasets.}
\label{tab:tab1}
\begin{tabular}{|l|c|c|c|c|c|c|}
\toprule
\multirow{2}{*}{}   & \multicolumn{3}{|c|}{D1}                                                                                                        & \multicolumn{3}{|c|}{D2}                                                                                                        \\
                    & \begin{tabular}[c]{@{}c@{}}Minority \\ class\end{tabular} & \begin{tabular}[c]{@{}c@{}}Majority \\ class\end{tabular} & Total & \begin{tabular}[c]{@{}c@{}}Minority \\ class\end{tabular} & \begin{tabular}[c]{@{}c@{}}Majority \\ class\end{tabular} & Total \\
\midrule
Number of tweets    & 200                                                       & 303                                                       & 1350  & 81                                                        & 148                                                       & 1651  \\
Word count          & 1814                                                      & 3060                                                      & 12901 & 673                                                       & 1444                                                      & 14059 \\
Unique words        & 1106                                                      & 1510                                                      & 5440  & 505                                                       & 793                                                       & 6177  \\
Average words per tweet & 9.07                                                      & 10.09                                                     & 9.55  & 8.31                                                      & 9.76                                                      & 8.51 \\ \hline
\end{tabular}
\end{table*}

\subsection{Features Representation}
The accuracy of a classification system depends on its representation of the problem. In particular, in the case of the tweet classification task it is necessary to transform the text of the tweet, which is a string of characters, into an appropriate representation for the learning classifier. Therefore, in this study we use three different representations:

Bag of Words (BoW) \cite{sousa} is one of the most popular technique used for natural language processing and information retrieval, which focus on the number of occurrences of words. This produces a vocabulary of the unique words occurring in all tweets and used as feature vectors representing whether each word in the vocabulary is exist or not. 

Term Frequency Inverse Document Frequency (TF–IDF) is a weighing scheme based on the combination of Term Frequency (TF) and Inverse Document Frequency (IDF). This is often used in Information Retrieval and Text Mining, which converts the tweet text into a matrix of integers generating a sparse matrix of the counts \cite{garreta}.

Word Embedding (WE) \cite{yang2018using,mikolov2013distributed} is a powerful technique that has been used successfully in these recent years. We define a feature vector space that is composed of unsupervised of word embedding vectors. A set of word embedding vectors is the representation of the ideal semantic space of words in real-valued vector space, hence the relationships between word vectors mirror the linguistic relationships of the words. Word Embedding vectors are a dense representation of the word meaning, thus each word is linked to a real-valued vector with a defined dimension. Word embedding can be generated using either static pre-trained word embedding models such as Word2vec \cite{mikolov2013distributed}, GloVe \cite{pennington2014glove} and fastText \cite{joulin2017bag} or using contextual pre-trained embedding models such as BERT \cite{bertdevlin2018} and RoBERTA \cite{liu2019roberta}. Sentence embedding is similar to word embedding, where each embedding is a low-dimensional vector that represents a sentence in a dense representation. There are different algorithms to create sentence embedding, with the same goal of creating similar embedding for similar sentence are doc2Vec \cite{quoc} Infersent \cite{alexis} and Sentence-BERT \cite{nils}.
\subsection{Model Description}

In this section, we describe our proposed topic detection models. First, in Section \ref{TML} we present the traditional machine learning systems we have used. Second, in Section \ref{DL}  we discuss the deep learning architectures we have employed. Finally, in Section \ref{TL}  we describe the transfer learning models we have proposed.
\subsubsection{Traditional machine learning}  \label{TML}
To classify topics we consider the two models most commonly used in tweet classification: the Multinomial Naïve Bayes and Logistic Regression, which predict classes based on a combination of scores for each feature.
Multinomial Naïve Bayes (MNB) \cite{mccallum1998} is one of the most well-known classifiers since it has been shown to be highly accurate and effective in text classification.  This considers the frequency of words to produce data distribution in a multinomial manner. For the hyper-parameter optimization, we try the following hyperparameter and values: ‘Alpha’=[0.01, 0.1,0.2,0.3,0.4, 0.5,0.6,0.7,0.8,0.9, 1], ‘fit\_prior’=[True, False]. After performing hyperparameter optimization, we use the default configuration of the classifier for both datasets.

Logistic Regression (LR) \cite{shalev} is a non-linear regression model, which can be used in order to estimate the probability of a binary response y, based on a set of $z$ independent features $F = [f_{1}...f_{z}]$ for which $y \| f$ follows the Bernoulli distribution with a $P(F)$ probability of success \cite{carreras2018link}. For the hyperparameter optimization we try the following hyperparameter and values: ‘C’ = [0.01, 0.1, 1, 10], ‘class\_weight’ = [None, balanced], ‘penalty’ = [l1, l2] and ‘solver’ = [lbfgs, liblinear]. After performing hyperparameter optimization we set the ‘C’ value to 10, the ‘solver’ value to liblinear, the ‘class\_weight’ value to None and the ‘penalty’ value to L2 for D1 and the default configuration of the classifier for D2.

To extract features for their inclusion in each classifier, we use in both classifiers the three types of text representation explained in the previous section: TF–IDF, BoW and word embedding.

\subsubsection{Deep Learning} \label{DL}

In this section, we describe the deep learning methods explored in this study. Particularly, we use two variants of neural networks, which are explained below:

Convolution Neural Network (CNN) is one of the most important and effective neural network models, which is an alternative way of feeding forward neural network. The different layers in CNN are sparsely connected by connecting a local region of an input layer by the neuron in the next layer. \cite{kim2014convolutional} is the first work that applied CNN to text classification, where words are converted into numerical values by converting words into word embedding. A matrix of 2-dimensional is formed from each text where each row corresponds to a word vector in that text. Then the various steps of CNN consist of a convolution layer, a pooling layer, and a fully connected layer. 

Recurrent Neural Networks (RNNs) \cite{mikolov2010} is another class of neural network to tackle the problem of sequential learning posed by the traditional neural network. The connections between nodes form a directed graph along a temporal sequence, which allow it to exhibit temporal dynamic behavior.  The most popular variants of RNN being Long Short-Term Memory (LSTM), introduced by \cite{hochreiter1997long} which is trained using backpropagation through time. This type of networks has memory blocks so they are capable of learning temporal sequences and their long-term dependencies. In contrast, Bidirectional Long Short-Term Memory (BiLSTM) networks have a two way flow of information as the sequence is trained by two LSTM networks, one in the forward and the other in the reverse direction \cite{schuster1997bidirectional}. 

In this case, as feature representations we use word embeddings. We specifically use the set of 300 dimensional pre-trained vectors of GloVe \cite{pennington2014glove}. 

\subsubsection{Transfer Learning} \label{TL}

Transfer learning is the process of adapting a pre-trained neural network to a new dataset through transfer the learned features. In other words, is a technique to enhance learning in a new task by transferring the knowledge from a learned task \cite{torrey2010transfer}. This is widely used to develop models to solve problems where the availability of large data is limited in order to train, and evaluate the systems.

Transformer is an attention mechanism, which learns contextual relationships between words in a text. This includes two distinct mechanisms: an encoder that reads the textual input and a decoder that outputs a prediction for the task \cite{vaswani2017attention}. In contrast to directional models that read the textual input sequentially (right-to-left or left-to-right), the encoder reads the whole sequence at once. This allows the model to learn the context of a given word based on the entire of its surroundings. \cite{vaswani2017attention} achieved an enhancement in the translation task with the use of the attention mechanism without relying on RNN, paving the way for further transformer architectures.

Bidirectional Encoder Representations from Transformers (BERT) is the first transformer-based language model introduced by Google. This auto-encoding language model trained using stacked encoder blocks from Transformers with a masked language modeling to learn embedding bidirectionally \cite{bertdevlin2018}. The model is pretrained on large unsupervised text corpora using two self-supervision tasks: Masked Language Modeling and Next Sentence Prediction. The first task is implemented by masking 15\% of the words randomly in every sentence and training the model to predict them. The second task is a classification task with two sentences input and the model is expected to recognize the original order between these two sentences, which increases the document level understanding. The success of BERT lead to the emergence of many variants, including RoBERTa which is a Robust BERT approach introduced by \cite{liu2019roberta}; DistilBert  which is a distilled BERT \cite{sanh2019distilbert}  and ALBERT which is a lite BERT \cite{lan2019albert} .

Efficiently Learning an Encoder that Classifies Token Replacements Accurately (ELECTRA), introduced as a collaboration between Stanford University and Google, which uses a discriminative pre-training approach. According to \cite{clark2020electra}, this method requires substantially less data and computational resources than RoBERTa to perform in similar capacity as the latter models. Analysis showed that when using only 1/4th of the data ELECTRA performs similar to RoBERTa but outperformed RoBERTa when using a similar amount of training data. The training approach presented is called replaced token detection, where the input sentences are corrupted similar to BERT. However, instead of masking tokens, here a token is replaced with a new token generated using a small generator network similar to a masked language model. The pre-training task focuses on classifying tokens into two classes, i.e., tokens that are replaced and tokens that remain unchanged.

Sentence Transformer (ST) is a popular approach used for semantic similarity, clustering and semantic search. ST encode a unique vector representation of a textual sentence based on its semantic signature. The representation is built through contrastive training by adapting a transformer model in a Siamese architecture \cite{nils}, targeting to maximize the distance between sentences that are semantically dissimilar, and minimize the distance between semantically similar sentences. The idea of using ST for text classification is not new while includes an encoding step and a classification step (e.g. Logistic Regression). To the best of our knowledge, we did not find any work that performed an end-to-end STF for tweet classification in a Siamese manner. The reason is that STs were pre-trained on semantic similarity tasks as well as it is intuitive to further adapt them for the same objective. In this paper, we explore pre-trained STs for tweet classification, particularly topic detection.

\section{Experiments and Evaluation} \label{ER}
In this section, we present the experimental and evaluation results we carried out to show the effectiveness of STF. Through the experiments on two benchmark Twitter dataset, we try to find answers to the following two questions:
\begin{itemize}
    \item \textbf{RQ1}: Can STF improve the accuracy of tweet topic classification?
    \item \textbf{RQ2}: Does STF need a large amount of labelled data in order to achieve a good performance?
\end{itemize}
To answer these questions, first, we introduce pre-processing step we have applied to the chosen datasets as well as the hyperparameter optimization for neural networks and transfer learning fine-tuning. Then, we present the performance results in terms of overall Accuracy.

\subsection{Tweet preprocessing}
Preprocessing is one of the key modules of natural language processing systems, especially in text classification \cite{uysal2014impact}. In fact, Twitter data have become a challenge for natural language processing tools that are usually designed for well written texts \cite{hutto2014vader}. They contain mentions, emojis, elongated words, hashtags, expressions and urls that make tokenization difficult. To overcome those challenges we follow these steps:
\begin{itemize}
    \item Annotating and unpacking hashtags splitting the hashtag to its constituent words (e.g., \#ILoveComputerScience becomes  $<$/hashtag$>$ I Love Computer Science $<$/hashtag$>$).
    \item Normalizing URLs, emails, users’ mentions, the word RT, percent, money, time, date expressions and phone numbers, stop words, special characters (emoticons) and numerical characters.
    \item Annotating and reducing elongated words (e.g. gooooal becomes  $<$/elongated$>$) goal) and repeat words (e.g.  go go go go   $<$/repeated$>$) go).
    \item Converting all tweets to lower case.
\end{itemize}
\subsection{Hyperparameter optimization for neural networks}
The neural networks leveraged in this study contain a number of hyperparameters that must be estimated to achieve optimal results. For this purpose, we perform 5-fold cross validation with the train dataset in order determine the best hyperparameters. Table \ref{tab:tab2} shows the hyperparameters options that have been tested on each dataset and the resulting best parameters for each model (LSTM, CNN and BiLSTM). To avoid overfitting during supervised training of a neural network, we use early stopping as a form of regularization by stopping training before the weights have converged.

% Please add the following required packages to your document preamble:
% \usepackage{multirow}
\begin{table*}[]
\centering
\caption{Best hyperparameter values selection of the Deep Learning models.}
\label{tab:tab2}
\begin{tabular}{|l|l|c|c|c|c|c|c|}
\toprule
\multirow{2}{*}{Parameter} & \multirow{2}{*}{Options}                                                   & \multicolumn{2}{|c|}{CNN} & \multicolumn{2}{|c|}{LSTM} & \multicolumn{2}{|c|}{BiLSTM} \\
                                &                                                                            & D1         & D2         & D1          & D2         & D1           & D2          \\
\midrule
Size                            & {[}50,75,100{]}                                                            & 50         & 100        & 75          & 50         & 75           & 50          \\
Dropout                         & {[}0.25, 0.5{]}                                                            & 0.25       & 0.25       & 0.25        & 0.5        & 0.5          & 0.25        \\
Activation                      & {[}tanh, relu{]}                                                           & relu       & tanh       & relu        & tanh       & relu         & relu        \\
Optimizer                       & {[}Adam, SGD{]}                                                            & Adam       & Adam       & Adam        & Adam       & Adam         & Adam        \\
Batch size                      & {[}8, 16, 32, 64{]}                                                        & 32         & 8          & 32          & 16         & 64           & 8           \\
Learning rate                   & \begin{tabular}[c]{@{}l@{}}{[}0.001, 0.002, \\ 0.01 , 0.02{]}\end{tabular} & 0.01       & 0.01       & 0.002       & 0.02       & 0.002        & 0.01 \\ \hline
\end{tabular}
\end{table*}
\subsection{Transfer learning fine-tuning}
The transformers bases models are pre-trained on general corpora, which it has learned general knowledge from formal textual data. However, in this work the downstream task is topic identification from tweets that means we are dealing with informal text. Therefore, it is important to analyze the contextual information extracted from the pre-trained layers of language models and then fine-tune it using our interested dataset. By fine-tuning we update weights using the annotated dataset that is new to the already trained model. For the optimizer, we use the Adam optimizer \cite{kingma2014adam,zhang2019adam} which performs well for natural language processing. For the purposes of fine-tuning, authors in \cite{sun2019fine} have recommended choosing from the following parameters values: learning rate, batch size, number of epoch and max sequence. We fine-tuned the models by trying the following hyperparameter and values : epoch=[2, 3, 4, 5, 10, 20], Batch\_size=[8, 16, 32] and learning rate [$1e^{-5}$, $2e^{-5}$, $3e^{-5}$, $4e^{-5}$, $5e^{-5}$]. The optimized hyperparameters of each model are presented in Table \ref{tab:tab3}. The max sequence length was established to 64 in all the experiments.

% Please add the following required packages to your document preamble:
% \usepackage{multirow}
\begin{table*}[]
\centering
\caption{Best hyperparameters values selection of pre-trained language models(transformers based and sentence transformers based).}
\label{tab:tab3}
\begin{tabular}{|p{5.9cm}|c|c|c|c|c|c|}
\toprule
\multirow{3}{*}{}                                                            & \multicolumn{2}{|c|}{Epoch}                   & \multicolumn{2}{|c|}{Batch size}        & \multicolumn{2}{|c|}{Learning rate}  \\
\midrule
                                                                                  & D1                   & D2                   & D1                 & D2               & D1                  & D2                 \\
\midrule
electra-base-discriminator                                                 & 20                   & 20                   & 8                  & 32               & 2e-5                & 3e-5               \\
electra-base-generator                                                     & 10                   & 20                   & 8                  & 8                & 2e-5                & 5e-5               \\
distilbert-base-uncased                                                           & 3                    & 4                    & 8                  & 8                & 2e-5                & 5e-5               \\
distilroberta-base                                                                & 5                    & 20                   & 8                  & 8                & 2e-5                & 2e-5               \\
bert-base-uncased                                                                 & 5                    & 20                   & 8                  & 16               & 2e-5                & 4e-5               \\
albert-base-v2                                                                    & 10                   & 10                   & 16                 & 8                & 2e-5                & 2e-5               \\
roberta-base                                                                      & 10                   & 10                   & 8                  & 16               & 1e-5                & 2e-5               \\
paraphrase-mpnet-base-v2-fuzzy-matcher & 20                   & 10                   & 8                  & 8                & 2e-5                & 2e-5               \\
paraphrase-mpnet-base-v2                                                          & 4                    & 5                    & 8                  & 8                & 2e-5                & 2e-5               \\
all-mpnet-base-v1                                                                 & 10                   & 10                   & 32                 & 8                & 4e-5                & 2e-5               \\
all-mpnet-base-v2                                                                 & 10                   & 5                    & 8                  & 8                & 2e-5                & 2e-5               \\
all-MiniLM-L12-v2                                                                 & 4                    & 10                   & 8                  & 8                & 2e-5                & 2e-5               \\
paraphrase-albert-small-v2                                                        & 3                    & 10                   & 8                  & 8                & 2e-5                & 2e-5               \\
all-roberta-large-v1                                                              & 2                    & 3                    & 8                  & 16               & 2e-5                & 2e-5               \\
paraphrase-MiniLM-L3-v2                                                           & 10                   & 20                   & 8                  & 16               & 2e-5                & 4e-5               \\
all-distilroberta-v1                                                              & 20                   & 20                   & 8                  & 8                & 2e-5                & 2e-5 \\ \hline
\end{tabular}
\end{table*}

\subsection{Results analysis}

In this section we explore the capabilities and the limits of the different approaches we have evaluated. To do so, we refer to the following parameters: 
\begin{itemize}
    \item True Positive (TP): the instances correctly classified as Positive. 
    \item True Negative (TN): the instances correctly classified as Negative. 
    \item False Positive (FP): the instances wrongly classified as Positive. 
    \item False Negative (FN): the instances wrongly classified as Negative. 
\end{itemize}
Using the aforementioned parameters, several measures can be drawn; we are interested in the overall Accuracy, which can be calculated as follows: 
\begin{equation}
\centering Accuracy = (TP + TN)/(TP + FP + TN + FN)
\end{equation}
In our experiments, we leveraged  Scikit-Learn \footnote{\cite{garreta} Scikit-learn (also known as sklearn) a machine learning library, which features various classification algorithms and features extractcion tecchniques.} , Ktrain \footnote{\cite{ktrain2020ktrain} Ktrain is a library to help build, train, debug, and deploy neural networks in the deep learning software framework with only a few lines of code} and Keras \footnote{Keras is an open source neural network library written in Python which is designed for fast experimentation with deep neural networks. \url{https://keras.io}} which uses Tensorflow \footnote{TensorFlow is an open-source software library for data flow programming across a range of tasks, which is used for different machine learning applications, such as neural networks. \url{https://www.tensorflow.org}} as back-end. We run all experiments on a machine with an Intel Core i7-7700 CPU (2.80GHz) and 16GB RAM.

We compared the performance of the different models on topic detection in two datasets. For that purpose, we used 5-fold cross validation. Table \ref{tab:tab4} shows the prediction performances we have achieved for each dataset, and each classifier.
In all the models, we leveraged word embedding as the input features, while in the case of traditional machine learning algorithms, also we tested the statistical-features BoW and TF–IDF including LR-BoW, LR-TF-IDF, MNB-BoW and MNB-TF–IDF. In most machine learning classifiers, BoW produces better results than the TF-IDF and word embeddings features, for both D1 and D2.
The baseline experiments (deep learning and traditional machine learning models) does not performed well due to the lack of sufficient training instances. In terms of the Accuracy score, the deep learning networks achieve better results than traditional algorithms in D1 dataset, but this is not the case for D2, where the best baseline is the MNB-BoW classifier. These findings are in line with the work of \cite{sousa} where traditional machine learning were found to have comparable accuracy results to deep neural networks on tweet topic classification task.

As shown in Table \ref{tab:tab4}, the pre-trained language model based on sentence transformers substantially outperforms the baselines systems for both datasets. The best performance was achieved by all-roberta-large-v1 followed by all-mpnet-base-v1 and all-mpnet-base-v1, which worked in the same way in both datasets. It is important to note that for both datasets; Sentence transformers based models achieve better results than transformers based models, yielded overall Accuracy results of 91.63\% and 77.52\% for D1 and D2, respectively.

% Please add the following required packages to your document preamble:
% \usepackage{multirow}
\begin{table*}[]
\centering
\caption{Accuracy results on D1 and D2.}
\label{tab:tab4}
\begin{tabular}{|l|l|l|l|}
\toprule
                                                                                         & Model                                                                             & D1    & D2    \\
\midrule
\multirow{6}{*}{\begin{tabular}[c]{@{}l@{}}Traditional \\ machine learning\end{tabular}} & MNB                                                                               & 63.38 & 53.12 \\
                                                                                         & LR                                                                                & 64.96 & 54.42 \\
                                                                                         & MNB-TF\_IDF                                                                       & 67.38 & 58.33 \\
                                                                                         & LR-TF\_IDF                                                                        & 68.88 & 58.73 \\
                                                                                         & MNB-BoW                                                                           & 69.28 & 59.13 \\
                                                                                         & LR-BoW                                                                            & 69.33 & 58.70 \\
\midrule
\multirow{3}{*}{Deep Learning}                                                           & CNN                                                                               & 76.69 & 48.23 \\
                                                                                         & LSTM                                                                              & 72.41 & 45.88 \\
                                                                                         & BiLSTM                                                                            & 73.76 & 47.00 \\
\midrule
\multirow{7}{*}{Transformers}                                                            & electra-base-discriminator                                                        & 85.48 & 72.75 \\
                                                                                         & electra-base-generator                                                            & 84.52 & 70.65 \\
                                                                                         & distilbert-base-uncased                                                           & 86.52 & 74.24 \\
                                                                                         & distilroberta-base                                                                & 85.85 & 72.69 \\
                                                                                         & bert-base-uncased                                                                 & 88.00 & 74.73 \\
                                                                                         & albert-base-v2                                                                    & 84.66 & 71.39 \\
                                                                                         & roberta-base                                                                      & 86.07 & 73.12 \\
\midrule
\multirow{9}{*}{Sentence Transformers}                                                   & \begin{tabular}[c]{@{}l@{}}paraphrase-mpnet-base-v2-fuzzy-matcher\end{tabular} & 88.44 & 75.78 \\
                                                                                         & paraphrase-mpnet-base-v2                                                          & 88.59 & 75.10 \\
                                                                                         & all-mpnet-base-v1                                                                 & 89.77 & 76.65 \\
                                                                                         & all-mpnet-base-v2                                                                 & 90.00 & 76.65 \\
                                                                                         & all-MiniLM-L12-v2                                                                 & 87.70 & 74.55 \\
                                                                                         & paraphrase-albert-small-v2                                                        & 85.70 & 71.57 \\
                                                                                         & all-roberta-large-v1                                                              & \textbf{91.63} & \textbf{77.52} \\
                                                                                         & paraphrase-MiniLM-L3-v2                                                           & 85.26 & 71.95 \\
                                                                                         & all-distilroberta-v1                                                              & 88.14 & 76.28\\ \hline
\end{tabular}
\end{table*}

Subsequently, we compare the performance of STF with four existing machine learning based approaches: (A) \cite{vadivukarassi}, which used TF–IDF and MNB; (B) \cite{sousa}, which used BoW, TF–IDF and LR; C \cite{hamoud}, which used BoW and SVM; D \cite{mello}, which used BoW and ensemble learning; and four existing deep learning based approaches, namely: (E) \cite{serban2019real}, which used Fasttext and CNN; (F) \cite{serban2019real}, which used Fasttext and LSTM; (G) \cite{oliveira2018politicians}, which used Word2vec based on Google News and CNN; (H) \cite{oliveira2018politicians}, which used Word2vec based on Google News and LSTM. 

While comparing our proposed STF with the other state-of-the-art classifiers in Table \ref{tab:tab5}, the Accuracy of our proposed STF is the highest. One reason could be that STF was able to capture long-term dependencies. When comparing with CNN and LSTM, STF does not requires a large amount of well-labeled training data to achieve good performance unlike to the most of deep learning approaches. When comparing with MNB and LR, STF has the benefit of not having to extract and design handcrafted features. While dealing with social media data, where the user tends to use lot of slangs, abbreviation etc. However, the language structure is not preserved; STF helps to process the input words while correlating with its surrounding ones. Although quite effective, current tweet topic detection approaches are costly, as they require a huge amount of labeled tweets in order to achieve good accuracy results. The tweets labeling process is very expensive and labor-intensive while hinders the deployment of artificial intelligence systems in the industry. In contrast, STF does not requires a huge amount of well-labeled training data to achieve a good performance, which makes it suitable in for both big and small data. Furthermore, previous state-of-the-art traditional machine learning classification approaches leverages handcrafted features. This type of approach has confronted to the curse of dimensionality and data sparseness. Conversely, STF has the ability to learn features from the textual data automatically. 

To sum up our findings, experimental results clearly show that our STF outperforms the existing classification models with a difference between [16.52\% and 37.19\%] and [18.76\% and 34.42\%] using D1 and D2, respectively. This yielded Accuracy results of 91.63\% and 77.52\% using D1 and D2, respectively. Comparing these results we highlight the significance of fine-tuning sentence transformers models, since that those models that achieves the best results in both datasets. Furthermore, we highlight the importance of hyperparameter tuning to find the best combination of model hyperparameters in each dataset. 

\begin{table}[]
\centering
\caption{State-of-the-art results for topic classification (* indicate the difference between STF and that baseline is statistically significant (t-test, p $>$ 0.05)).}
\label{tab:tab5}
\begin{tabular}{|c|c|c|}
\toprule
Model & D1 & D2 \\
\midrule
A*                          & 67.38                   & 58.33                   \\
B*                          & 68.44                   & 58.76                   \\
C*                          & 54.44                   & 43.22                   \\
D*                          & 69.78                   & 58.39                   \\
E*                          & 74.59                   & 46.25                   \\
F*                          & 70.96                   & 43.10                   \\
G*                          & 75.11                   & 46.69                   \\
H*                          & 71.41                   & 43.84                   \\
STF                         & \textbf{91.63}                   & \textbf{77.52}  \\ \hline         
\end{tabular}
\end{table}

\section{Conclusion} \label{CC}
The approach taken STF leveraged pre-trained sentence transformers and fine-tuning, yet it yields high accuracy results. Our proposition highlights a new area in tweet topic classification approaches based on transfer learning. The evaluation of STF was done using two benchmark datasets. This will allow researchers to compare different systems fairly against our system. Experiments show that: (1) the proposed STF improves the accuracy of topic classification and outperforms existing state-of-the-art approaches; (2) the proposed STF does not require a huge amount of labeled data in order to achieve good performance. In future works, we plan to pursue several directions. First, we want to focus on the sentence transformer model and try to adjust its vocabulary to support the tweet classification task. A costlier technique could be to consider training a novel sentence transformers model that is customized for tweet classification tasks. Second, we plan to test and compare different data augmentation approaches in order to deal with the problem of small-sample learning. We plan to create a more advanced classifier architecture on top of the sentence transformers model that can give better results than a simple feed-forward neural network. From a research standpoint, we will use STF to analyze Twitter conversations in different contexts to determine the extent of topical discussions with public discourse, and to understand how their sophistication and capabilities evolve over time.

\end{document}